\pdfoutput=1

\documentclass[11pt]{article}

\usepackage{acl}

\usepackage{times}
\usepackage{latexsym}

\usepackage[T1]{fontenc}

\usepackage[utf8]{inputenc}

\usepackage{microtype}
\usepackage{graphicx}
\usepackage{xspace}
\usepackage{colortbl}

\usepackage{tabularx}
\usepackage{booktabs}
\usepackage{multirow}
\usepackage{comment}
\usepackage{caption}
\usepackage{subcaption}
\usepackage{linguex}
\renewcommand{\eachwordone}{\itshape}
\usepackage{textgreek}
\usepackage{amssymb}
\usepackage{amsmath}
\usepackage{newunicodechar}
\usepackage{tipa}
\usepackage[T4,T1]{fontenc}
\usepackage{siunitx}
\usepackage{pifont}

\newcommand*{\yoruba}{Yor\`ub\'a\xspace}
\newcommand*{\ghomala}{Ghom\'al\'a'\xspace}
\newcommand*{\ewe}{\'Ew\'e\xspace}
\newcommand*{\zulu}{isiZulu\xspace}
\newcommand*{\xhosa}{isiXhosa\xspace}
\newcommand*{\shona}{chiShona\xspace}
\newcommand*{\swahili}{Kiswahili\xspace}
\newcommand*{\mafand}{MAFAND-MT\xspace}

\newenvironment{tfour}{\fontencoding{T4}\selectfont}{}

\newcount\Comments  
\usepackage{color}
\newcommand{\kibitz}[2]{\ifnum\Comments=1\textcolor{#1}{#2}\fi}

\definecolor{Gray}{gray}{0.85}
\definecolor{LightCyan}{rgb}{0.88,1,1}
\newcolumntype{a}{>{\columncolor{Gray}}r}
\newcolumntype{b}{>{\columncolor{LightCyan}}r}

\newcommand*\samethanks[1][\value{footnote}]{\footnotemark[#1]}
\title{MasakhaPOS: Part-of-Speech Tagging for Typologically Diverse African Languages}

\author{\normalsize Cheikh M. Bamba Dione$^{1, \dagger,}$\thanks{ \ \  Equal contribution. }, \ David Ifeoluwa Adelani$^{2,\dagger,}$\samethanks, \ \ Peter Nabende$^{3,\dagger}$,  Jesujoba O. Alabi$^{4,\dagger}$,  \\
\textbf{\normalsize Thapelo Sindane$^{5}$, Happy Buzaaba$^{6\dagger}$, Shamsuddeen Hassan Muhammad$^{7, 8\dagger}$, } \\
\textbf{ \normalsize Chris Chinenye Emezue$^{9,10\dagger}$,  Perez Ogayo$^{11\dagger}$, Anuoluwapo Aremu$^{\dagger}$, Catherine Gitau$^{\dagger}$, }\\
\textbf{\normalsize Derguene Mbaye$^{12\dagger}$, 
 Jonathan Mukiibi$^{3\dagger}$, Blessing Sibanda$^{\dagger}$, Bonaventure F. P. Dossou$^{10,13,14\dagger}$, } \\
\textbf{\normalsize  Andiswa Bukula$^{15}$, Rooweither Mabuya$^{15}$,  Allahsera Auguste Tapo$^{16\dagger}$, Edwin Munkoh-Buabeng$^{17\dagger}$, } \\
\textbf{\normalsize Victoire Memdjokam Koagne$^{\dagger}$, Fatoumata Ouoba Kabore$^{18\dagger}$, Amelia Taylor$^{19}$, Godson Kalipe$^{\dagger}$,} \\
\textbf{\normalsize Tebogo Macucwa$^{5}$,  Vukosi Marivate$^{5,13\dagger}$, Tajuddeen Gwadabe$^{\dagger}$, Elvis Tchiaze Mboning$^{\dagger}$, } \\
\textbf{\normalsize  Ikechukwu Onyenwe$^{20}$,  Gratien Atindogbe$^{21}$, Tolulope Anu Adelani$^{\dagger}$, Idris Akinade$^{22}$, } \\ 
\textbf{\normalsize  Olanrewaju Samuel$^{\dagger}$,  Marien Nahimana,  Théogène Musabeyezu, Emile Niyomutabazi,  } \\ 
\textbf{\normalsize Ester Chimhenga, Kudzai Gotosa, Patrick Mizha, Apelete Agbolo$^{23}$, Seydou Traore$^{24}$, } \\
\textbf{\normalsize Chinedu Uchechukwu$^{20}$, Aliyu Yusuf$^{8}$, Muhammad Abdullahi$^{8}$, Dietrich Klakow$^{4}$ } \\\\
\footnotesize
$^\dagger$Masakhane NLP, $^1$Université Gaston Berger, Senegal,
 $^2$University College London, UK, $^3$Makerere University, Uganda, \\
\footnotesize
 $^4$Saarland University, Germany, $^{5}$University of Pretoria, South Africa,
$^6$ RIKEN Center for AIP, Japan, \\
\footnotesize
$^{7}$Bayero University Kano, Nigeria. 
$^{8}$University of Porto, Portugal,
$^{9}$Technical University of Munich, Germany,
$^{10}$Lanfrica, \\
\footnotesize
$^{11}$Carnegie Mellon University, USA,
$^{12}$Baamtu, Senegal,
$^{13}$Lelapa AI,
$^{14}$Mila Quebec AI Institute, Canada,\\
\footnotesize
$^{15}$SADiLaR, South Africa, 
$^{16}$Rochester Institute of Technology, USA, 
$^{17}$TU Clausthal, Germany, 
$^{18}$Uppsala University, Sweden, 
\\
\footnotesize
$^{19}$Malawi University of Business and Applied Science, Malawi, 
$^{20}$Nnamdi Azikiwe University, Nigeria, 
\\
\footnotesize
$^{21}$University of Buea, Cameroon,
$^{22}$University of Ibadan, Nigeria,
$^{23}$Ewegbe Akademi, Togo,
$^{24}$AMALAN, Mali.
\\
}

\begin{document}
\maketitle
\begin{abstract}
In this paper, we present MasakhaPOS, the largest part-of-speech (POS) dataset for 20 typologically diverse African languages. We discuss the challenges in annotating POS for these languages using the UD (universal dependencies) guidelines. We conducted extensive POS baseline experiments using conditional random field and several multilingual 
pre-trained language models. We applied various cross-lingual transfer models trained with data available in UD. Evaluating on the MasakhaPOS dataset, we show that choosing the best transfer language(s) in both single-source and multi-source setups greatly improves the POS tagging performance of the target languages, in particular when combined with cross-lingual parameter-efficient fine-tuning methods. Crucially, transferring knowledge from a language that matches the language family and morphosyntactic properties seems more effective for POS tagging in unseen languages.
\end{abstract}

\section{Introduction}

Part-of-Speech (POS) tagging is a process of assigning the most probable grammatical category (or tag) to each word (or token) in a given sentence of a particular natural language. POS tagging is one of the fundamental steps for many natural language processing (NLP) applications, including machine translation, parsing, text chunking, spell and grammar checking. While great strides have been made for (major) Indo-European languages such as English, French and German, work on the African languages is quite scarce. 
The vast majority of African languages lack annotated datasets for training and evaluating basic NLP systems. 

There have been recent works on the development of benchmark datasets for training and evaluating models
in African languages for various NLP tasks, including machine translation ~\cite{team2022NoLL,adelani-etal-2022-thousand}, text-to-speech~\cite{ogayo22_interspeech,meyer22c_interspeech}, speech recognition~\cite{Ritchie2022LargeVS}, sentiment analysis ~\cite{muhammad-etal-2022-naijasenti,muhammad2023afrisenti}, news topic classification~\citep{adelani2023masakhanews}, and named entity recognition~\cite{adelani-etal-2021-masakhaner,adelani-etal-2022-masakhaner}. However, there is no large-scale dataset for POS covering several African languages. 

To tackle the data bottleneck issue for low-resource languages, recent work applied cross-lingual transfer \cite{artetxe-etal-2020-cross, pfeiffer-etal-2020-mad,ponti-etal-2020-xcopa}
using multilingual pretrained language models (PLMs) \cite{conneau-etal-2020-unsupervised} to model specific phenomena in low-resource target languages.
While such a cross-lingual transfer is often evaluated by fine-tuning multilingual models on English data, more recent work has shown that English is not often the best transfer language \cite{lin-etal-2019-choosing,de-vries-etal-2022-make,adelani-etal-2022-masakhaner}.  

\vspace{-1mm}
\paragraph{Contributions} In this paper, we develop \textbf{MasakhaPOS}~ --- the largest POS dataset for 20 typologically diverse African languages.
We highlight the challenges of annotating POS for these diverse languages using the universal dependencies (UD)~\citep{nivre2016universal} guidelines such as tokenization issues, and POS tags ambiguities. We provide extensive POS baselines using conditional random field (CRF) and several multilingual pre-trained language models (PLMs). Furthermore, we experimented with different parameter-efficient cross-lingual transfer methods~\cite{pfeiffer-etal-2021-unks,ansell-etal-2022-composable}, and transfer languages with available training data in the UD. Our evaluation 
demonstrates that choosing the best
transfer language(s) in both single-source and multi-source setups leads to large improvements in POS tagging
performance, especially when combined with parameter-fine-tuning methods. Finally, we show that a transfer language that belongs to the same language family and shares similar morphological characteristics (e.g. Non-Bantu Niger-Congo) seems to be more effective for tagging POS in unseen languages. 
For reproducibility, we release our code, data and models on GitHub\footnote{\url{https://github.com/masakhane-io/masakhane-pos}}


\section{Related Work}
In the past, efforts have been made to build a POS tagger for several African languages, including Hausa \cite{tukur2020parts}, Igbo \cite{onyenwe2014part}, Kinyarwanda \cite{cardenas2019grounded}, Luo \cite{de2010knowledge},    Setswana \cite{malema2017setswana,malema2020complex}, \xhosa \cite{delman2016development}, Wolof \cite{dione2010design}, \yoruba \cite{adedjouma2012part, ishola-zeman-2020-yoruba}, and \zulu \cite{koleva2013towards}. While POS tagging has been investigated for the aforementioned languages, annotated datasets exist only in a few African languages. In the Universal dependencies dataset~\cite{nivre2016universal},  nine African languages\footnote{including Amharic, Bambara, Beja, \yoruba, and Zaar with no training data in UD.} are represented. Still, only four of the nine languages have training data, i.e. Afrikaans, Coptic, Nigerian-Pidgin, and Wolof. 
In this work, we create the largest POS dataset for 20 African languages following the UD annotation guidelines. 



\section{Languages and their characteristics}
We focus on 20 Sub-Saharan African languages, spoken in circa 27 countries in the Western, Eastern, Central and Southern regions of Africa.
An overview of the focus languages is provided in \autoref{tab:languages}. 
The selected languages represent four language families: Niger-Congo (17), Afro-Asiatic (Hausa), Nilo-Saharan (Luo), and English Creole (Naija). 
Among the Niger-Congo languages, eight belong to the Bantu languages.

\begin{table*}[th!]
 \footnotesize
 \begin{center}
 \resizebox{\textwidth}{!}{%
  \begin{tabular}{lllr|llrr}
    \toprule
     & &\textbf{African} & \textbf{No. of}  &  \multicolumn{2}{c}{\textbf{}}   & \textbf{\#} & \textbf{Average sentence} \\
    \textbf{Language} & \textbf{Family} & \textbf{Region} & \textbf{Speakers}   & \textbf{Source} & \textbf{Train / dev / test}    & \textbf{Tokens} & \textbf{Length (\# Tokens)} \\
    \midrule
    Bambara (\texttt{bam}) & NC / Mande & West & 14M & \mafand~\cite{adelani-etal-2022-thousand} & 793/ 158/ 634 & 40,137 &  25.9 \\
    \ghomala (\texttt{bbj}) & NC / Grassfields &Central& 1M & \mafand & 750/ 149/ 599 & 23,111 & 15.4 \\
    \ewe (\texttt{ewe}) & NC / Kwa &West& 7M  & \mafand & 728/ 145/ 582 & 28,159 & 19.4 \\
    Fon (\texttt{fon}) & NC / Volta-Niger &West& 2M & \mafand  & 798/ 159/ 637 & 49,460 & 30.6 \\
    Hausa (\texttt{hau}) & Afro-Asiatic / Chadic &West& 63M & Kano Focus and Freedom Radio & 753/ 150/ 601 & 41,346 & 27.5\\
    Igbo (\texttt{ibo}) & NC / Volta-Niger &West& 27M  & IgboRadio and Ka \d{O}d\d{I} Taa  & 803/ 160/ 642  & 52,195 & 32.5 \\
    Kinyarwanda (\texttt{kin}) & NC / Bantu &East& 10M & IGIHE, Rwanda & 757/ 151/ 604  & 40,558 & 26.8\\
    Luganda (\texttt{lug}) & NC / Bantu &East& 7M & \mafand &  733/ 146/ 586 & 24,658 & 16.8\\
    Luo (\texttt{luo}) & Nilo-Saharan &East& 4M  & \mafand &  757/ 151/ 604 & 45,734 & 30.2\\
    Mossi (\texttt{mos}) & NC / Gur &West& 8M & \mafand &  757/ 151/ 604 & 33,791 & 22.3 \\
    Chichewa (\texttt{nya}) & NC / Bantu &South-East& 14M & Nation Online Malawi & 728/ 145/ 582  & 24,163 & 16.6 \\
    Naija (\texttt{pcm}) & English-Creole &West& 75M & \mafand & 752/ 150/ 600  &  38,570 & 25.7\\
    \shona (\texttt{sna}) & NC / Bantu &South& 12M & VOA Shona & 747/ 149/ 596  & 39,785 & 26.7\\
    Kiswahili (\texttt{swa}) & NC / Bantu &East \& Central & 98M & VOA Swahili & 675/ 134/ 539 & 40,789 & 29.5\\
    Setswana (\texttt{tsn}) & NC / Bantu &South& 14M & \mafand &  753/ 150/ 602 & 41,811 & 27.9 \\
    Akan/Twi (\texttt{twi}) & NC / Kwa &West& 9M & \mafand &  775/ 154/ 618 & 41,203 & 26.2 \\
    Wolof (\texttt{wol}) & NC / Senegambia &West& 5M & \mafand & 770/ 154/ 616  & 44,002 & 28.2 \\
    \xhosa (\texttt{xho}) & NC / Bantu &South& 9M & Isolezwe Newspaper & 752/ 150/ 601  & 25,313 & 16.8 \\
    \yoruba (\texttt{yor}) & NC / Volta-Niger &West& 42M & Voice of Nigeria and Asejere & 875/ 174/ 698  & 43,601 & 24.4\\
    \zulu (\texttt{zul}) & NC / Bantu &South& 27M & Isolezwe Newspaper & 753/ 150/ 601 & 24,028 & 16.0 \\
    \bottomrule
  \end{tabular}
  }
  \caption{\textbf{Languages and Data Splits for MasakhaPOS Corpus}. Language, family (NC: Niger-Congo), number of speakers, news source, and data split in number of sentences.}
  \vspace{-4mm}
  \label{tab:languages}
  \end{center}
\end{table*}

The writing system of our focus languages is mostly based on Latin script (sometimes with additional letters and diacritics). 
Besides Naija, \swahili, and Wolof, the remaining languages are all tonal. As far as morphosyntax is concerned, noun classification is a prominent grammatical feature for an important part of our focus languages. 12 of the languages \textit{actively} make use of between 6--20 noun classes. This includes all Bantu languages, \ghomala, Mossi, Akan and Wolof~\cite{Van_de_Velde2006-fz,Payne2017,Bodomo2002TheMO,BabouLoporcaro+2016+1+57}. 
Noun classes can play a central role in POS annotation. 
For instance, in \xhosa, adding the class prefix can change the grammatical category of the word \cite{delman2016development}. 
All languages use the SVO word order, while Bambara additionally uses the SOV word order. 
\autoref{sec:appendix_lang_char} provides the details about the language characteristics. 

\section{Data and Annotation for MasakhaPOS}
\subsection{Data collection}
\autoref{tab:languages} provides the data source used for POS annotation --- collected from online newspapers. The choice of the news domain is threefold. First, it is the second most available resource after the religious domain for most African languages. Second, it covers a diverse range of topics. Third, the news domain is one of the dominant domains in the UD. We collected \textbf{monolingual news corpus} with an open license for about eight African languages, mostly from local newspapers. For the remaining 12 languages, we make use of MAFAND-MT~\cite{adelani-etal-2022-thousand} \textbf{translation corpus} that is based on the news domain. While there are a few issues with translation corpus such as translationese effect,  we did not observe serious issues in annotation. The only issue we experienced was a few misspellings of words, which led to annotators labeling a few words with the "X" tag. However, as a post-processing step, we corrected the misspellings and assigned the correct POS tags. 

\subsection{POS Annotation Methodology}

For the POS annotation task, we collected \textbf{1,500 sentences per language}. 
As manual POS annotation is very tedious, we agreed to manually annotate 100 sentences per language
in the first instance. This data is then used as training data for automatic POS tagging (i.e., fine-tuning RemBERT~\cite{chung2021rethinking} PLM) of the remaining unannotated sentences. Annotators proceeded to fix the mistakes of the predictions (i.e. 1,400 sentences). This drastically reduced the manual annotation efforts since a few tags are predicted with almost 100\% accuracy like punctuation marks, numbers and symbols. Proper nouns were also predicted with high accuracy due to the casing feature. 

To support work on manual corrections of annotations, 
most of the languages used the IO Annotator\footnote{\href{{https://ioannotator.com/}}{https://ioannotator.com/} }
 tool, a collaborative annotation platform for text and images.
The tool provides support for multi-user annotations simultaneously on datasets. 
For each language, 
we hired three native speakers with linguistics backgrounds to perform POS annotation.\footnote{Each annotator was paid \$750 for 1,500 sentences.} 
To ensure high-quality annotation, we recruited a language coordinator to supervise annotation in each language. 
In addition, we provided online support (documentation and video tutorials) to
train annotators on POS annotation. 
We made use of the Universal POS tagset \cite{petrov-etal-2012-universal}, which contains 17 tags.\footnote{\href{{https://universaldependencies.org/u/pos/}}{https://universaldependencies.org/u/pos/} }
To avoid the use of spurious tags, for each word to be annotated, annotators have to choose 
one of the possible tags made available on the IO Annotator tool through a drop-down menu.
For each language, annotation was done independently by each annotator. 
At the end of annotation, language coordinators worked with their team to resolve disagreements using IOAnnotator or Google Spreadsheet. We refer to our newly annotated POS dataset as \textbf{MasakhaPOS}. 


\subsection{Quality Control}

Computation of automatic inter-agreement metrics scores like Fleiss Kappa was a bit challenging due to tokenization issues, e.g. many compound family names are split. Instead, we adopted the tokenization defined by annotators since they are annotating all words in the sentence. Due to several annotation challenges as described in \autoref{sec:challenges}, seven language teams (\ghomala, Fon, Igbo, Chichewa \shona, \swahili, and Wolof) decided to engage annotators on online calls (or in person discussions) to agree on the correct annotation for each word in the sentence. The other language teams allowed their annotators to work individually, and only discuss sentences on which they did not agree. Seven of the 13 languages achieved a sentence-level annotation agreement of over $75\%$.  Two more languages (Luganda and \zulu) have sentence-level agreement scores of between $64.0\%$ to $67.0\%$. The remaining four languages (Ewe, Luo, Mossi, and Setswana) only agreed on less than $50\%$ of the annotated sentences. 
This confirms the difficulty of the annotation task for many language teams.  Despite this challenge, we ensured that all teams resolved all disagreements to produce high-quality POS corpus. \autoref{sec:annotation_agreement} provides details of the number of agreed annotation by each language team. 

After quality control, we divided the annotated sentences into training, development and test splits consisting of 50\%, 10\%, 40\% of the data respectively. We chose a larger test set proportion that is similar to the size of test sets in the UD, usually larger than 500 sentences.  \autoref{tab:languages} provides the details of the data split. We split very long sentences into two to fit the maximum sequence length of 200 for PLM fine-tuning. We further performed manual checks to correct sentences split at arbitrary parts. 

\section{Annotation challenges}
\label{sec:challenges}
When annotating our focus languages, we faced two main challenges: tokenization and POS ambiguities.




\subsection{Tokenization and word segmentation}
In UD, the basic annotation units are syntactic words (rather than phonological or orthographical words) \cite{nivre2016universal}. Accordingly, clitics need to be split off and contraction must be undone where necessary. Applying the UD annotation scheme to our focus languages was not straightforward due to the nature of those languages, especially with respect to the notion of word, the use of clitics and multiword units.

\subsubsection{Definition of word}
For many of our focus languages (e.g.\ Chichewa, Luo, \shona, Wolof and \xhosa), it was difficult to establish a dividing line between a word and a phrase. For instance, the \shona word \emph{ndakazomuona} translates into English as a whole sentence (`I eventually saw him'). This word consists of several morphemes that convey distinct morphosyntactic information \cite{chabata2000shona}: \emph{Nda-} (subject concord), \emph{-ka-} (aspect), \emph{-zo-} (auxiliary), \emph{-mu-} (object concord), \emph{-ona-} (verb stem). This illustrates pronoun incorporation \cite{bresnan1987topic}, i.e. subject and/or object pronouns appear as bits of morphology on a verb or other head, functioning as agreement markers. 
 Naturally, one may want to split this word into several tokens reflecting the different grammatical functions. For UD, however, morphological features such as agreement are encoded as properties of words and there is no attempt at segmenting words into morphemes, implying that items like \emph{ndakazomuona} should be treated as a single unit. 

\subsubsection{Clitics} \label{sec-clit}
In languages like Hausa, Igbo, IsiZulu, Kinyarwanda, Wolof and \yoruba, we observed an extensive use of cliticization. Function words such as prepositions, conjunctions, auxiliaries and determiners can attach to other function or content words. For example, the Igbo contracted form \emph{yana} consists of a pronoun (PRON) \emph{ya} and a coordinating conjunction (CCONJ) \emph{na}. Following UD, we segmented such contracted forms, as they correspond to multiple (syntactic) words. However, there were many cases of fusion where a word has morphemes that are not necessarily easily segmentable. For instance, the \shona word \emph{vave} translates into English as `who (PRON) are (AUX) now (ADV)'. Here, the morpheme \emph{-ve}, which functions both as auxiliary and adverb, cannot be further segmented, even though it corresponds to multiple syntactic words. Ultimately, we treated the word \emph{vave} as a unit, which received the AUX POS tag. 


In addition, there were word contractions with phonological changes, posing serious challenges, as proper segmentation may require to recover the underlying form first. For instance, the Wolof contracted form ``cib" \cite{dione2019developing} consists of the preposition \textit{ci} `in' and the indefinite article \textit{ab} `a'. However, as a result of phonological change, the initial vowel of the article is deleted. Accordingly, to properly segment the contracted form, it won't be sufficient to just extract the preposition \textit{ci} because the remaining form \textit{b} will not have meaning. 
Also, some word contractions are ambiguous. For instance, in Wolof, a form like \textit{geek} can be split into \textit{gi} ‘the’ and \textit{ak} where \textit{ak} can function as a conjunction ‘and’ or as a preposition ‘with’.

\subsubsection{One unit or multitoken words?}
Unlike the issue just described in \ref{sec-clit}, it was sometimes necessary to go in the other direction, and combine several orthographic tokens into a single syntactic word. Examples of such multitoken words are found e.g.\ in Setswana \cite{malema2017setswana}. For instance, in the relative structure \emph{ngwana yo o ratang} (the child who likes ...), the relative marker \emph{yo o} is a multitoken word that matches the noun class (class 1) of the relativized noun \emph{ngwana} (`child'), which is subject of the verb \emph{ratang} (`to like'). 
In UD, multitoken words are allowed for a restricted class of phenomena, such as numerical expressions like 20 000 and abbreviations (e. g.). We advocate that this restricted class be expanded to phenomena like Setswana relative markers. 

\subsection{POS ambiguities}
There were cases where a word form lies on the boundary between two (or more) POS categories. 

\subsubsection{Verb or conjunction?}
In quite a few of our focus languages (e.g.\ \yoruba, Wolof), 
a form of the verb `say' is also used as a subordinate conjunction (to mark out clause boundaries) with verbs of speaking. For example, in the \yoruba sentence \emph{Olú gbàgbé \textbf{pé} Bolá tí jàde} (lit.\ `Olu forgot that Bola has gone') \cite{lawal1991yoruba}, the item \emph{pé} seems to behave both like a verb and a subordinate conjunction. 
On the one hand, because of the presence of another verb \emph{gbàgbé} `to forget', the pattern may be analyzed as a serial verb construction (SVC) \cite{oyelaran1982scope,Gueldemann-2008}, i.e. a construction that contains sequences of two or more verbs without any syntactic marker of subordination. This would mean that \emph{pé} is a verb. On the other hand, however, this item shows properties of a complementizer \cite{lawal1991yoruba}. For instance, \emph{pé} can occur in sentence initial position, which in \yoruba is typically occupied by subordinating conjunctions. Also, unlike verbs, \emph{pé} cannot undergo reduplication for nominalization (an ability that all \yoruba verbs have). This seems to provide evidence for treating this item as a subordinate conjunction rather than a verb.

\subsubsection{Adjective or Verb?}
In some of our focus languages, the category of adjectives is not entirely distinct morpho-syntactically from verbs. In Wolof and \yoruba, the notions that would be expressed by adjectives in English are encoded through verbs \cite{mclaughlin2004there}. Igbo \citep{welmers2018african} and \ewe \cite{mclaughlin2004there} have a very limited set of underived adjectives (8 and 5, respectively). For instance, in Wolof, unlike in English, an `adjective' like \emph{gaaw} `be quick' does not need a copula (e.g.\ `be' in English) to function as a predicate. Likewise, the Bambara item \emph{téli} `quick' as in the sentence \emph{Sò ka téli} `The horse is quick' \cite{aplonova2017towards} has adjectival properties, as it is typically used to modify nouns and specify their properties or attributes. It also has verbal properties, as it can be used in the main predicative position functioning as a verb. This is signaled by the presence of the auxiliary \emph{ka}, which is a special predicative marker \emph{ka} that typically accompanies qualitative verbs \cite{vydrin2018corpus}.

\subsubsection{Adverbs or particles?}
The distinction between adverbs and particles was not always straightforward. For instance, many of our focus languages have ideophones, i.e.\ words that convey an idea by means of a sound (often reduplicated) that expresses an action, quality, manner, etc. Ideophones may behave like adverbs by modifying verbs for such categories as time, place, direction or manner. However, they can also function as verbal particles. For instance, in Wolof, an ideophone like \emph{jërr} as in \emph{tàng jërr} ``very hot'' (\emph{tàng} means ``to be hot'') is an intensifier that only co-occurs as a particle of that verb. Thus, it would not be motivated to treat it as another POS other than PART. Whether such ideophones are PART or ADV or the like varies depending on the language.


\section{Baseline Experiments}
\begin{table*}[t]
\begin{center}
\footnotesize
\resizebox{\textwidth}{!}{%
\begin{tabular}{lrrrrrrrrrrrrrrrrrrrr|c}
\toprule
\textbf{Model} & \textbf{bam} & \textbf{bbj} & \textbf{ewe} & \textbf{fon} & \textbf{hau} & \textbf{ibo} & \textbf{kin} & \textbf{lug} & \textbf{luo} & \textbf{mos} & \textbf{nya} & \textbf{pcm} & \textbf{sna} & \textbf{swa} & \textbf{tsn} & \textbf{twi}  & \textbf{wol} & \textbf{xho} & \textbf{yor} & \textbf{zul} & \textbf{AVG}   \\
\midrule
CRF & 89.1 & 78.9 & 88.0 & 88.1 & 89.8 & 75.2 & 95.3 & 88.3 & 84.6 & 86.0 & 77.7 & 85.6 & 85.9 & 89.3 & 81.4 &  81.5 & 91.0 & 81.8 & 92.0 & 84.2 & 85.7\\
\midrule
\multicolumn{7}{l}{\textit{Massively-multilingual PLMs}} \\
mBERT (172M) & 89.9 & 75.2 & 86.0 & 87.6 & 90.7 & 76.5 & 96.9 & 89.6 & 87.0 & 86.5 & 79.9 & 90.4 & 87.5 & 92.0 & 81.9 & 83.9 & 92.5 & 85.9 & 93.4 & 86.8 & $87.0$ \\ 
XLM-R-base (270M) & 90.1 & 83.6 & 88.5 & 90.1 & 92.5 & 77.2 & 96.7 & 89.1 & 87.2 & 90.7 & 79.9 & 90.5 & 87.9 & 92.9 & 81.3 & 84.1 & 92.4 & 87.4 & 93.7 & 88.0 & $88.2$ \\
XLM-R-large (550M) & 90.2 & \textbf{85.4} & 88.8 & 90.2 & 92.8 & 78.1 & 97.3 & 90.0 & 88.0 & 91.1 & 80.5 & 90.8 & 88.1 & \textbf{93.2} & 82.2 & 84.9 & \textbf{92.9} & 88.1 & 94.2 & 89.4 &  $88.8$ \\ 
RemBERT (575M) & 90.6 & 82.6 & \textbf{88.9} & \textbf{90.8} & \textbf{93.0} & \textbf{79.3} & 98.0 & 90.3 & 87.5 & 90.4 & 82.4 & 90.9 & 89.1 & 93.1 & 83.6 & \textbf{86.0} & 92.1 & \textbf{89.3} & 94.7 & \textbf{90.2} & $89.1$ \\ 
\midrule
\multicolumn{7}{l}{\textit{Africa-centric PLMs}} \\
AfroLM (270M)& 89.2  & 77.8 & 87.5 & 82.4 & 92.7 & 77.8 & 97.4 & 90.8 & 86.8 & 89.6 & 81.1 & 89.5 & 88.7 & 92.8 & \textbf{83.8} & 83.9 & 92.1 & 87.5 & 91.1 & 88.8 & $87.6$ \\ 
AfriBERTa-large (126M) & 89.4 & 79.6 & 87.4 & 88.4 & \textbf{93.0} & \textbf{79.3} & 97.8 & 89.8 & 86.5 & 89.9& 79.7 & 89.8 & 87.8 & 93.0 & 82.5 & 83.7 & 91.7 & 86.1 & 94.5 & 86.9 & $87.8$ \\ 
AfroXLMR-base (270M) & 90.2 & 83.5 & 88.5 & 90.1 & \textbf{93.0} & 79.1 & 98.2 & 90.9 & 86.9 & 90.9 & 82.7 & 90.8 & 89.2 & 92.9 & 82.7 & 84.3 & 92.4 & 88.5 & 94.5 & 89.4 & $88.9$ \\ 
AfroXLMR-large (550M) & \textbf{90.5} & 85.3 &88.7 & 90.4 & \textbf{93.0} & 78.9 & \textbf{98.4} & \textbf{91.6} & \textbf{88.1} & \textbf{91.2} & \textbf{83.2} & \textbf{91.2} & \textbf{89.5} &  \textbf{93.2} & 83.0 & 84.9  & \textbf{92.9} & 88.7 & \textbf{95.0} & 90.1 & $\mathbf{89.4}$ \\ 
\bottomrule
    \end{tabular}
    }
\caption{\textbf{Accuracy of baseline models on MasakhaPOS dataset }. We compare several multilingual PLMs including the ones trained on African languages. Average is over 5 runs. }
\label{tab:pos_baselines}
  \end{center}
\end{table*}

\begin{table*}[!ht]
    \centering
    \footnotesize
    \resizebox{\textwidth}{!}{%
    \begin{tabular}{lrrrrrrraarraarrar||r}
    \toprule
        ~ & \textbf{ADJ} & \textbf{ADP} & \textbf{ADV} & \textbf{AUX} & \textbf{CCONJ} & \textbf{DET} & \textbf{INTJ} & \textbf{NOUN} & \textbf{NUM} & \textbf{PART} & \textbf{PRON} & \textbf{PROPN} & \textbf{PUNCT} & \textbf{SCONJ} & \textbf{SYM} & \textbf{VERB} & \textbf{X} & \textbf{ACC} \\ 
        \midrule
        \textbf{bam} & 41.0 & 77.0 & 72.0 & 82.0 & 91.0 & 0.0 & ~ & 91.0 & 90.0 & 95.0 & 97.0 & 82.0 & 100.0 & 71.0 & 25.0 & 83.0 & 0.0 & 90.7 \\ 
        \textbf{bbj} & 71.0 & 80.0 & 67.0 & 89.0 & 84.0 & 85.0 & 0.0 & 82.0 & 86.0 & 78.0 & 91.0 & 92.0 & 100.0 & 88.0 & ~ & 86.0 & ~ & 85.6 \\ 
        \textbf{ewe} & 72.0 & 83.0 & 57.0 & ~ & 94.0 & 89.0 & 100.0 & 91.0 & 91.0 & 87.0 & 90.0 & 93.0 & 100.0 & 84.0 & 13.0 & 82.0 & ~ & 88.7 \\
        \textbf{fon} & 91.0 & 88.0 & 69.0 & 75.0 & 94.0 & 96.0 & ~ & 91.0 & 90.0 & 89.0 & 95.0 & 91.0 & 100.0 & 51.0 & ~ & 89.0 & ~ & 90.4 \\ 
        \textbf{hau} & 86.0 & 80.0 & 71.0 & 96.0 & 89.0 & 84.0 & 0.0 & 94.0 & 98.0 & 95.0 & 76.0 & 98.0 & 99.0 & 86.0 & ~ & 96.0 & 62.0 & 92.9 \\ 
        \textbf{ibo} & 95.0 & 89.0 & 56.0 & 98.0 & 76.0 & 79.0 & 0.0 & 70.0 & 95.0 & 0.0 & 98.0 & 95.0 & 100.0 & 6.0 & 0.0 & 81.0 & ~ & 79.2 \\ 
        \textbf{kin} & 86.0 & 99.0 & 91.0 & 0.0 & 100.0 & 99.0 & ~ & 99.0 & 100.0 & 84.0 & 98.0 & 97.0 & 100.0 & 97.0 & 0.0 & 99.0 & 0.0 & 98.4 \\ 
        \textbf{lug} & 71.0 & 96.0 & 72.0 & 90.0 & 90.0 & 76.0 & ~ & 94.0 & 93.0 & 94.0 & 15.0 & 94.0 & 100.0 & 89.0 & ~ & 92.0 & ~ & 91.6 \\ 
        \textbf{luo} & 73.0 & 88.0 & 69.0 & 87.0 & 69.0 & 82.0 & ~ & 89.0 & 96.0 & 86.0 & 42.0 & 89.0 & 100.0 & 94.0 & 100.0 & 86.0 & 0.0 & 88.2 \\ 
        \textbf{mos} & 64.0 & 83.0 & 72.0 & 91.0 & 93.0 & 84.0 & ~ & 91.0 & 93.0 & 94.0 & 83.0 & 90.0 & 100.0 & 95.0 & ~ & 92.0 & ~ & 91.2 \\ 
        \textbf{nya} & 74.0 & 79.0 & 56.0 & 25.0 & 77.0 & 81.0 & 20.0 & 92.0 & 86.0 & 12.0 & 73.0 & 86.0 & 99.0 & 6.0 & ~ & 89.0 & ~ & 83.1 \\ 
        \textbf{pcm} & 78.0 & 97.0 & 74.0 & 86.0 & 98.0 & 92.0 & ~ & 95.0 & 98.0 & 90.0 & 86.0 & 91.0 & 98.0 & 86.0 & 45.0 & 91.0 & ~ & 91.1 \\ 
        \textbf{sna} & 51.0 & 94.0 & 44.0 & 87.0 & 89.0 & 83.0 & ~ & 95.0 & 96.0 & 0.0 & 78.0 & 92.0 & 99.0 & 58.0 & 60.0 & 94.0 & ~ & 89.4 \\ 
        \textbf{swa} & 95.0 & 86.0 & 65.0 & 82.0 & 95.0 & 56.0 & ~ & 97.0 & 98.0 & 86.0 & 51.0 & 97.0 & 100.0 & 91.0 & ~ & 95.0 & 0.0 & 93.1 \\ 
        \textbf{tsn} & 57.0 & 80.0 & 82.0 & 42.0 & 53.0 & 78.0 & 17.0 & 94.0 & 97.0 & 62.0 & 76.0 & 91.0 & 99.0 & 18.0 & 0.0 & 95.0 & 0.0 & 82.4 \\ 
        \textbf{twi} & 55.0 & 82.0 & 68.0 & 52.0 & 87.0 & 93.0 & 0.0 & 86.0 & 77.0 & 21.0 & 82.0 & 92.0 & 100.0 & 9.0 & 0.0 & 87.0 & ~ & 84.8 \\ 
        \textbf{wol} & 0.0 & 94.0 & 81.0 & 94.0 & 96.0 & 90.0 & 22.0 & 91.0 & 90.0 & 98.0 & 92.0 & 96.0 & 100.0 & 85.0 & 62.0 & 94.0 & ~ & 92.9 \\ 
        \textbf{xho} & 73.0 & 69.0 & 47.0 & 17.0 & 88.0 & 54.0 & 0.0 & 87.0 & 100.0 & ~ & 80.0 & 95.0 & 100.0 & 57.0 & 0.0 & 90.0 & ~ & 88.3 \\ 
        \textbf{yor} & 84.0 & 92.0 & 82.0 & 99.0 & 97.0 & 97.0 & ~ & 95.0 & 94.0 & 83.0 & 95.0 & 96.0 & 100.0 & 98.0 & ~ & 95.0 & 0.0 & 95.1 \\ 
        \textbf{zul} & 68.0 & 26.0 & 72.0 & 21.0 & 67.0 & 82.0 & 0.0 & 91.0 & 99.0 & ~ & 81.0 & 99.0 & 100.0 & 91.0 & 100.0 & 91.0 & 96.0 & 90.0 \\ 
        \midrule
        \textbf{AVE} &  69.2 & 83.1 & 68.4 & 69.1 & 86.4 & 79.0 & 15.9 & 90.8 & 93.4 & 69.7 & 79.0 & 92.8 & 99.7 & 68.0 & 33.8 & 90.4 & 19.8 & 89.4 \\ 
        \bottomrule
    \end{tabular}
    }
    \caption{\textbf{Tag distribution of the ``AfroXLMR-large'' -based POS tagger } (reporting results from the first run). The tags with high average accuracy ($>90.0 \%$ ) across all languages are highlighted in \colorbox{Gray}{gray}. }
    \label{tab:pos_tag_dist}
\end{table*}

\subsection{Baseline models}
We provide POS tagging baselines using both CRF  and multilingual PLMs. For the PLMs, we fine-tune three massively multilingual PLMs pre-trained on at least 100 languages (mBERT~\cite{devlin-etal-2019-bert}, XLM-R~\cite{conneau-etal-2020-unsupervised}, and RemBERT~\cite{chung2021rethinking}), and three Africa-centric PLMs like AfriBERTa~\cite{ogueji-etal-2021-small}, AfroXLMR~\cite{alabi-etal-2022-adapting}, and AfroLM~\cite{Dossou2022AfroLMAS} pre-trained on several African languages. The baseline models are:

\paragraph{CRF} is one of the most successful sequence labeling approach prior to PLMs. CRF models the sequence labeling task as an undirected graphical model, using both labelled observations and contextual information as features. We implemented the CRF model using \texttt{sklearn-crfsuite},\footnote{ \href{{https://sklearn-crfsuite.readthedocs.io/}}{https://sklearn-crfsuite.readthedocs.io/}
} using the following features: the word to be tagged, two consecutive previous and next words, the word in lowercase, prefixes and suffixes of words, length of the word, and other boolean features like is the word a digit, a punctuation mark, the beginning of a sentence or end of a sentence. 

\paragraph{Massively multilingual PLM} We fine-tune mBERT, XLM-R (base \& large), and RemBERT pre-trained on 100-110 languages, but only few African languages. mBERT, XLM-R, and RemBERT were pre-trained on two (\texttt{swa} \& \texttt{yor}), three (\texttt{hau}, \texttt{swa}, \& \texttt{xho}), and eight (\texttt{hau}, \texttt{ibo}, \texttt{nya}, \texttt{sna}, \texttt{swa}, \texttt{xho}, \texttt{yor}, \& \texttt{zul}) of our focus languages respectively. The three models were all pre-trained using masked language model (MLM), mBERT and RemBERT additionally use the next-sentence prediction objective. 

\paragraph{Africa-centric PLMs} We fine-tune AfriBERTa, AfroLM and AfroXLMR (base \& large). The first two PLMs were pre-trained using XLM-R style pre-training, AfroLM additionally make use of active learning during pre-training to address data scarcity of many African languages. On the other hand, AfroXLMR was created through language adaptation~\citep{pfeiffer-etal-2020-mad} of XLM-R on 17 African languages, ``eng'', ``fra'', and ``ara''. AfroLM was pre-trained on all our focus languages, while AfriBERTa and AfroXLMR were pre-trained on 6 (\texttt{hau}, \texttt{ibo}, \texttt{kin}, \texttt{pcm}, \texttt{swa}, \& \texttt{yor}) and 10 (\texttt{hau}, \texttt{ibo}, \texttt{kin}, \texttt{nya}, \texttt{pcm}, \texttt{sna}, \texttt{swa}, \texttt{xho}, \texttt{yor}, \& \texttt{zul}) respectively. We fine-tune all PLMs using the HuggingFace Transformers library~\cite{wolf-etal-2020-transformers}. 

For PLM fine-tuning, we make use of a maximum sequence length of $200$,
batch size of $16$, gradient accumulation of $2$, learning rate of $5e-5$, and number of epochs 50. The
experiments were performed
on using Nvidia V100 GPU.



\subsection{Baseline results}
\autoref{tab:pos_baselines} shows the results of training POS taggers for each focus language using the CRF and PLMs. Suprinsingly, the CRF model gave a very impressive result for all languages with only a few points below the best PLM ($-3.7$). In general, fine-tuning PLMs gave a better result for all languages. The mBERT performance is ($+1.3$) better in accuracy than CRF. AfroLM and AfriBERTa are only slightly better than mBERT with ($<1$ point). One of the reasons for AfriBERTa's poor performance is that most of the languages are unseen during pre-training.\footnote{14 out of 20 languages are unseen} On the other hand, AfroLM was pre-trained on all our focus languages but on a small dataset (0.73GB) which makes it difficult to train a good representation for each of the languages covered during pre-training. 
Furthermore, XLM-R-base gave slightly better accuracy on average than both AfroLM ($+0.6$) and AfriBERTa ($+0.4$) despite seeing fewer African languages. However, the performance of the AfroXLMR-base exceeds that of XLM-R-base because it has been further adapted to 17 typologically diverse African languages, and the performance ($\pm 0.1$) is similar to the larger PLMs i.e RemBERT and XLM-R-large. 

Impressive performance was achieved by large versions of massively multilingual PLMs like XLM-R-large and RemBERT, and AfroXLMR (base \& large) i.e  better than mBERT ($+1.8$ to $+2.4$) and better than CRF ($+3.1$ to $+3.7$). 
The performance of the large PLMs (e.g. AfroXLMR-large) is larger for some languages when compared to mBERT like \texttt{bbj} ($+10.1$), \texttt{mos} ($+4.7$), \texttt{nya} ($+3.3$), and \texttt{zul} ($+3.3$). Overall, AfroXLMR-large achieves the best accuracy on average over all languages ($89.4$) because it has been pre-trained on more African languages with larger monolingual data and it's large size. Interestingly, 11 out of 20 languages reach an impressive accuracy of ($>90\%$) with the best PLM which is an indication of consistent and high quality POS annotation. 

\paragraph{Accuracy by tag distribution} \autoref{tab:pos_tag_dist} shows the POS tagging results by tag distribution using our best model ``AfroXLMR-large''. The tags that are easiest (with accuracy over $>90\%$) to detect across all languages are \texttt{PUNCT}, \texttt{NUM}, \texttt{PROPN}, \texttt{NOUN}, and \texttt{VERB}, while the most difficult are \texttt{SYM}, \texttt{INTJ}, and \texttt{X} tags. The difficult tags are often infrequent, which does not affect the overall accuracy. Surprisingly, a few languages like \yoruba and Kinyarwanda,  have very good accuracy on almost all tags except for the infrequent tags in the language.

\section{Cross-lingual Transfer}
\begin{figure*}[t]
    \centering
    \includegraphics[width=0.90\linewidth]{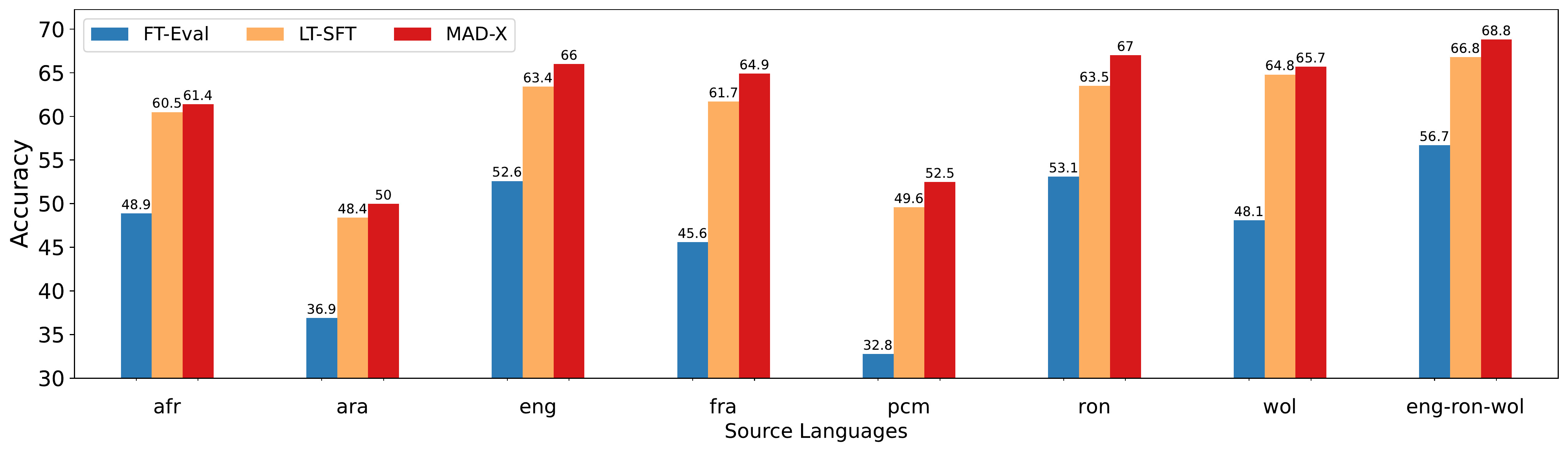}
    \vspace{-2mm}
    \caption{\textbf{Zero-shot cross-lingual transfer results using FT-Eval, LT-SFT and MAD-X}. Average over 20 languages. Experiments performed using AfroXLMR-base. Evaluation metric is Accuracy. }
    \label{fig:ave_transfer_plot}
    \vspace{-1mm}
    
\end{figure*}

\subsection{Experimental setup for effective transfer}
The effectiveness of zero-shot cross-lingual transfer depends on several factors including the choice of the best performing PLM, choice of an effective cross-lingual transfer method, and the choice of the best source language for transfer. Oftentimes, the source language chosen for cross-lingual transfer is English due to the availability of training data which may not be ideal for distant languages especially for POS tagging~\citep{de-vries-etal-2022-make}. 
To further improve performance, parameter-efficient fine-tuning approaches~\citep{pfeiffer-etal-2020-mad,ansell-etal-2022-composable} can be leveraged with additional monolingual data for both source and target languages. We highlight how we combine these different factors for effective transfer below: 

\paragraph{Choice of source languages}
Prior work on the choice of source language for POS tagging shows that the most important features are geographical similarity, genetic similarity (or closeness in language family tree) and word overlap between source and target language~\cite{lin-etal-2019-choosing}. 
We choose seven source languages for zero-shot transfer based on the following criteria (1) \textbf{availability of POS training} data in UD,\footnote{\url{https://universaldependencies.org/}}. Only three African languages satisfies this criteria (Wolof, Nigerian-Pidgin, and Afrikaans) (2) \textbf{geographical proximity} to African languages -- this includes non-indigeneous languages that have official status in Africa like English, French, Afrikaans, and Arabic. (3) \textbf{language family similarity} to target languages. The languages chosen are: \textit{Afrikaans} (\texttt{afr}), \textit{Arabic} (\texttt{ara}), \textit{English} (\texttt{eng}), \textit{French} (\texttt{fra}), \textit{Nigerian-Pidgin} (\texttt{pcm}), \textit{Wolof} (\texttt{wol}), and \textit{Romanian} (\texttt{ron}). While Romanian does not satisfy the last two criteria - it was selected based on the findings of \citet{de-vries-etal-2022-make} --- Romanian achieves the best transfer performance to the most number of languages in UD. \autoref{sec:ud_pos_data} shows the data split for the source languages.

\paragraph{Parameter-efficient cross-lingual transfer}The standard way of zero-shot cross-lingual transfer involves \textit{fine-tuning} a multilingual PLM on the source language labelled data (e.g. on a POS task), and \textit{evaluate} it on a target language. We refer to it as \textbf{FT-Eval} (or Fine-tune \& evaluate). However, the performance is often poor for unseen languages in PLM and distant languages. One way to address this is to perform  language adaptation using monolingual corpus in the target language before fine-tuning on the downstream task~\citep{pfeiffer-etal-2020-mad}, but this setup does not scale to many languages since it requires modifying all the parameters of the PLM and requires large disk space~\citep{alabi-etal-2022-adapting}. Several parameter-efficient approaches have been proposed like Adapters~\citep{pmlr-v97-houlsby19a} and Lottery-Ticketing Sparse Fine-tunings (LT-SFT)~\cite{ansell-etal-2022-composable} ---they are also modular and composable making them ideal for cross-lingual transfer. 

Here, we make use of \textbf{MAD-X 2.0}\footnote{an extension of MAD-X where
the last adapter layers are dropped, which has been shown to improve performance} adapter based approach~\cite{pfeiffer-etal-2020-mad,pfeiffer-etal-2021-unks} and \textbf{LT-SFT} approach. 
The setup is as follows: (1) We train language adapters/SFTs using monolingual news corpora of our focus languages. We perform language adaptation on the \textit{news} corpus to match the POS task domain, similar to \cite{alabi-etal-2022-adapting}. We provide details of the monolingual corpus in \autoref{sec:news_corpora}. (2) We train a task adapter/SFT on the source language labelled data using source language adapter/SFT. (3) We substitute the source language adapter/SFT with the target language/SFT to run prediction on the target language test set, while retaining the task adapter.

\paragraph{Choice of PLM} We make use of \textbf{AfroXLMR-base} as the backbone PLM for all experiments because it gave an impressive performance in \autoref{tab:pos_baselines}, and the availability of language adapters/SFTs for some of the languages by prior works~\cite{pfeiffer-etal-2021-unks,ansell-etal-2022-composable,alabi-etal-2022-adapting}. 
When a target language adapter/SFT of AfroXLMR-base is absent, XLM-R-base language adapter/SFT can be used instead since they share the same architecture and number of parameters, as demonstrated in \citet{alabi-etal-2022-adapting}.  
We did not find XLM-R-large based adapters and SFTs online,\footnote{\url{https://adapterhub.ml/}} and they are time-consuming to train especially for high-resource languages like English. 


\subsection{Experimental Results}

\begin{table*}[t]
\begin{center}
\footnotesize
\resizebox{\textwidth}{!}{%
\begin{tabular}{lraaararrrarrrrraarar|cc}
\toprule
\textbf{Method} & \textbf{bam} & \textbf{bbj} & \textbf{ewe} & \textbf{fon} & \textbf{hau} & \textbf{ibo} & \textbf{kin} & \textbf{lug} & \textbf{luo} & \textbf{mos} & \textbf{nya} & \textbf{pcm} & \textbf{sna} & \textbf{swa} & \textbf{tsn} & \textbf{twi}  & \textbf{wol} & \textbf{xho} & \textbf{yor} & \textbf{zul} & \textbf{AVG} & \textbf{AVG*}  \\
\midrule
\multicolumn{7}{l}{\texttt{\textbf{eng}} \textbf{as a source language}} \\
FT-Eval & 52.1 & 31.9 & 47.8 & 32.5 & 67.1 & 74.5 & 63.9 & 57.8 & 38.4 & 45.3 & 59.0 & 82.1 & 63.7 & 56.9 & 49.4 & 35.9 & 35.9 & 45.9 & 63.3 & 48.8 & $52.6$ & $51.9$ \\
LT-SFT & \textbf{67.9} & 57.6 & 67.9 & 55.5 & 69.0 & 76.3 & 64.2 & 61.0 & 74.5 & 70.3 & 59.4 & 82.4 & 64.6 & 56.9 & 49.5 & 52.1 & 78.2 & 45.9 & 65.3 & 49.8 & $63.4$ & $61.5$\\
MAD-X & 62.9 & 58.5 & 68.7 & 55.8 & 67.0 & 77.8 & 70.9 & 65.7 & 73.0 & 71.8 & \textbf{70.1} & 83.2 & 69.8 & 61.2 & 49.8 & 53.0 & 75.2 & \textbf{57.1} & 66.9 & \textbf{60.9} & $66.0$ & $64.5$ \\ 
\midrule
\multicolumn{7}{l}{\texttt{\textbf{ron}} \textbf{as a source language}} \\
FT-Eval & 46.5 & 30.5 & 37.6 & 30.9 & 67.3 & 77.7 & 73.3 & 56.9 & 36.7 & 40.6 & 62.2 & 78.9 & 66.3 & 61.0 & 55.8 & 35.7 & 33.8 & 49.6 & 63.5 & 56.3 & 53.1 & 52.7 \\
LT-SFT & 60.6 & 57.0 & 64.9 & 60.4 & 67.5 & 77.4 & 68.2 & 58.5 & 70.2 & 67.9 & 58.2 & 78.1 & 64.6 & 59.7 & 57.4 & 55.7 & 81.9 & 46.3 & 64.8 & 51.2 & $63.5$ & $61.7$ \\ 
MAD-X & 63.5 & 62.2 & 66.6 & 61.8 & 66.5 & 80.0 & \textbf{73.5} & 62.7 & 76.5 & 71.8 & 66.0 & 83.7 & 71.1 & \textbf{64.5} & \textbf{61.2} & 53.5 & 79.5 & 48.6 & 69.5 & 57.8 & $67.0$ & $65.4$  \\ 
\midrule
\multicolumn{7}{l}{\texttt{\textbf{wol}} \textbf{as a source language}} \\
FT-Eval & 40.8 & 36.5 & 39.8 & 37.4 & 55.1 & 58.6 & 49.2 & 51.8 & 35.1 & 44.9 & 49.0 & 51.6 & 53.8 & 42.9 & 45.0 & 38.4 & 88.6 & 46.0 & 52.5 & 45.5 & 48.1 & 45.7 \\
LT-SFT (N) & 64.4 & 64.3 & 69.8 & 63.0 & 67.0 & 79.7 & 63.7 & 64.0 & 74.1 & 72.2 & 56.5 & 72.7 & 67.7 & 53.0 & 51.3 & 56.2 & 92.5 & 46.0 & 69.8 & 47.7 & $64.8$ & $62.8$ \\ 
MAD-X (N) & 46.6 & 41.8 & 47.2 & 37.8 & 53.9 & 51.8 & 41.0 & 39.0 & 46.5 & 44.0 & 38.3 & 40.2 & 44.3 & 38.8 & 44.6 & 40.1 & 85.6 & 39.2 & 46.4 & 36.0 & $45.2$ 
 & $43.2$ \\ 
MAD-X (N+W) & 61.7 & 63.6 & 68.9 & 63.1 & 66.8 & 77.0 & 67.8 & \textbf{69.1} & 73.7 & 71.3 & 63.2 & 75.1 & 68.9 & 55.8 & 50.7 & 54.9 & 90.4 & 49.6 & 70.0 & 51.7 & $65.7$ & $63.8$ \\ 
\midrule
\multicolumn{7}{l}{\texttt{\textbf{multi-source: eng-ron-wol}}} \\
FT-Eval & 44.2 & 36.3 & 39.3 & 39.3 & 69.4 & 78.5 & 70.6 & 59.2 & 35.5 & 46.8 & 60.9 & 81.4 & 65.8 & 58.5 & 53.8 & 38.8 & 89.1 & 48.8 & 65.2 & 53.5 & 56.7 & 53.6 \\
LT-SFT & 67.4 & 64.6 & 70.0 & 64.2 & \textbf{70.4} & 81.1 & 68.7 & 63.9 & 76.4 & 73.9 & 58.8 & 83.0 & \textbf{69.6} & 57.3 & 52.7 & \textbf{57.2} & \textbf{93.1} & 45.8 & 69.8 & 48.3 & $66.8$ & $64.4$ \\ 
MAD-X &66.2 & \textbf{65.5} & \textbf{70.3} & \textbf{64.9} & 69.1 & \textbf{82.3} & 73.1 & 68.0 & \textbf{75.1} & \textbf{74.2} & 69.2 & \textbf{83.9} & 69.4 & 62.6 & 53.6 & 55.2 & 90.1 & 52.3 & \textbf{70.8} & 59.4 & $\mathbf{68.8}$ & $\mathbf{66.7}$ \\ 

\bottomrule
\vspace{-5mm}
    \end{tabular}
}
\caption{\textbf{Cross-lingual transfer to MasakhaPOS }. Zero-shot Evaluation using FT-Eval, LT-SFT, and MAD-X, with \texttt{ron}, \texttt{eng}, and \texttt{wol} as source languages. Experiments are based on AfroXLMR-base. Non-Bantu Niger-Congo languages highlighted with \colorbox{Gray}{gray}. AVG* excludes \texttt{pcm} and \texttt{wol} from the average since they are source languages. }
\label{tab:cross_lingual_transfer}
  \end{center}
  \vspace{-3mm}
\end{table*}

\paragraph{Parameter-efficient fine-tuning are more effective} \autoref{fig:ave_transfer_plot} shows the result of cross-lingual transfer from seven source languages with POS training data in UD, and their average accuracy on 20 African languages. We report the performance of the standard zero-shot cross-lingual transfer with AfroXLMR-base (i.e. FT-Eval), and parameter-efficient fine-tuning approaches i.e MAD-X and LT-SFT. Our result shows that MAD-X and LT-SFT gives significantly better results than FT-Eval, the performance difference is over 10\% accuracy on all languages. This shows the effectiveness of parameter-efficient fine-tuning approaches on cross-lingual transfer for low-resource languages despite only using small monolingual data (433KB - 50.2MB, as shown in \autoref{sec:news_corpora}) for training target language adapters and SFTs. Furthermore, we find MAD-X to be slightly better than LT-SFT especially when  \texttt{ron} ($+3.5$), \texttt{fra} ($+3.2$), \texttt{pcm} ($+2.9$), and \texttt{eng} ($+2.6$)  are used as source languages. 

\paragraph{The best source language} In general, we find \texttt{eng}, \texttt{ron}, and \texttt{wol} to be better as source languages to the 20 African languages. For the FT-Eval,  \texttt{eng} and \texttt{ron} have similar performance. However, for LT-SFT,  \texttt{wol} was slightly better than the other two, probably because we are transfering from an African language that shares the same family or geographical location to the target languages. For MAD-X, \texttt{eng} was surprisingly the best choice. 

\paragraph{Multi-source fine-tuning leads to further gains} \autoref{tab:cross_lingual_transfer} shows that co-training the best three source languages (\texttt{eng}, \texttt{ron}, and \texttt{wol}) leads to improved performance, reaching an impressive accuracy of $68.8\%$ with MAD-X. For the FT-Eval, we performed multi-task training on the combined training set of the three languages. LT-SFT supports multi-source fine-tuning --- where a task SFT can be trained on data from several languages jointly. However, MAD-X implementation does not support multi-source fine-tuning. We created our version of multi-source fine-tuning following these steps: (1) We combine all the training data of the three languages (2) We train a task adapter using the combined data and one of the best source languages' adapter. We experiment using \texttt{eng}, \texttt{ron}, and \texttt{wol} as source language adapter for the combined data. 
Our experiment shows that \texttt{eng} or \texttt{wol}  achieves similar performance when used as language adapter for multi-source fine-tuning. We only added the result using \texttt{wol} as source adapter on \autoref{tab:cross_lingual_transfer}. 
Appendix \autoref{sec:madx_multisource} provides more details on MAD-X multi-source fine-tuning. 

\paragraph{Performance difference by language family} 
\autoref{tab:cross_lingual_transfer} shows the transfer result per language for the three best source languages. \texttt{wol} has a better transfer performance to non-Bantu Niger-Congo languages in West Africa than \texttt{eng} and \texttt{ron},  especially for \texttt{bbj}, \texttt{ewe}, \texttt{fon}, \texttt{ibo}, \texttt{mos}, \texttt{twi}, and \texttt{yor} despite having a smaller POS training data (1.2k sentences) compared to \texttt{ron} (8k sentences) and \texttt{eng} (12.5k sentences). Also, \texttt{wol} adapter was trained on a small monolingual corpus (5.2MB). This result aligns with prior studies that choosing a source language from the same family leads to more effective transfer~\cite{lin-etal-2019-choosing,de-vries-etal-2022-make}. However, we find MAD-X to be more sensitive to the size of monolingual corpus. We obtained a very terrible transfer accuracy when we only train language adapter for \texttt{wol} on the news domain (2.5MB) i.e MAD-X (N), lower than FT-Eval. By additionally combining the news corpus with Wikipedia corpus (2.7MB) i.e MAD-X (N+W), we were able to obtain an impressive result comparable to LT-SFT. This highlight the importance of using larger monolingual corpus to train source language adapter. \texttt{wol} was not the best source language for Bantu languages probably because of the difference in language characteristics. For example,  Bantu languages are very morphologically-rich while non-Bantu Niger-Congo languages (like \texttt{wol}) are not.  Our further analysis shows that \texttt{sna} was better in transferring to Bantu languages. \autoref{sec:cross_lingual_all} provides result for the other source languages. 

\section{Conclusion}
In this paper, we 
created MasakhaPOS, the largest POS dataset for 20 typologically-diverse  
African languages. 
We showed that POS annotation of these languages based on the UD scheme can be quite challenging, especially 
with regard to word segmentation and POS ambiguities. 
We provide POS baseline models using CRF and by fine-tuning 
multilingual PLMs. We analyze cross-lingual transfer on MasakhaPOS dataset in single-source and multi-source settings. 
An important finding that emerged from this study is that choosing the appropriate transfer languages substantially improves POS tagging for unseen languages.  
The transfer performance is particularly effective when pre-training includes a language that shares typological features with the target languages.

\section{Limitations}

\paragraph{Some Language families in Africa not covered} For example,  Khoisan and Austronesian (like Malagasy). We performed extensive analysis and experiments on Niger-Congo languages but we only covered one language each in the Afro-asiatic (Hausa) and Nilo-Saharan (Dholuo) families. 

\paragraph{News domain} Our annotated dataset belong to the news domain, which is a popular domain in UD. However, the POS dataset and models may not generalize to other domains like speech transcript, conversation data etc. 

\paragraph{Transfer results may not generalize to all NLP tasks} We have only experimented with POS task, the best transfer language e.g for non-Bantu Niger-Congo languages i.e Wolof, may not be the same for other NLP tasks.

\section{Ethics Statement or Broader Impact}

Our work aims to understand linguistic characteristics of African languages, we do not see any potential harms when using our POS datasets and models to train ML models, the annotated dataset is based on the news domain, and the articles are publicly available, and we believe the dataset and POS annotation is unlikely to cause unintended harm. 

Also, we do not see any privacy risks in using our dataset and models  because it is based on news domain.

\section*{Acknowledgements}
This work was carried out with support from Lacuna Fund, an initiative co-founded by The Rockefeller Foundation, Google.org, and Canada’s International Development Research Centre. We are grateful to Sascha Heyer, for extending the ioAnnotator tool to meet our requirements for POS annotation. We appreciate the early advice from Graham Neubig, 
Kim Gerdes, and Sylvain Kahane on this project. David Adelani acknowledges the support of DeepMind Academic Fellowship programme. We appreciate all the POS annotators that contributed to this dataset. Finally, we thank
the Masakhane leadership, Melissa Omino, Davor
Orlic and Knowledge4All for their administrative ˇ
support throughout the project.
\bibliography{anthology,custom}
\bibliographystyle{acl_natbib}

\appendix


 \begin{table*}[t]
 \footnotesize
 \begin{center}
 \resizebox{\textwidth}{!}{%
  \begin{tabular}{lrlp{50mm}llllll}
    \toprule
     & \textbf{No. of} &\textbf{Latin Letters} &  &  &  &  & \textbf{Morphological} & \textbf{Inflectional} & \textbf{Noun}\\
    \textbf{Language} & \textbf{Letters} & \textbf{Omitted} & \textbf{Letters added}  & \textbf{Tonality} & \textbf{diacritics} & \textbf{Word Order} & \textbf{typology} & \textbf{Morphology (WALS)} & \textbf{Classes} \\
    \midrule
    Bambara (bam) & 27 & q,v,x & \textepsilon, \textopeno, \textltailn, \textipa{\ng} &  yes, 2 tones & yes & SVO \& SOV & isolating & strong suffixing & absent \\
    \ghomala (bbj) & 40 & q, w, x, y & bv, dz, \textschwa, a\textschwa, \textepsilon, gh, ny, nt, \textipa{\ng}, \textipa{\ng}k, \textopeno, pf, mpf, sh, ts, \textbaru, zh, '  &  yes, 5 tones & yes & SVO & agglutinative & strong prefixing & active, 6 \\
    \ewe (ewe) & 35  & c, j, q & \textrtaild, dz, \textepsilon, \textflorin, gb, \textgamma, kp, ny, \textipa{\ng}, \textopeno, ts, \textscriptv   & yes, 3 tones  & yes & SVO & isolating & equal prefixing and suffixing & vestigial \\
    Fon (fon) & 33 & q & \textrtaild, \textepsilon,gb, hw, kp, ny, \textopeno, xw & yes, 3 tones & yes  & SVO & isolating & little affixation & vestigial  \\
    Hausa (hau) & 44 & p,q,v,x & \texthtb, \texthtd, \texthtk, \begin{tfour}\m{y}\end{tfour}, kw, {\texthtk}w, gw, ky, {\texthtk}y, gy, sh, ts  & yes, 2 tones  & no & SVO & agglutinative & little affixation & absent \\
    Igbo (ibo) & 34 & c, q, x & ch, gb, gh, gw, kp, kw, nw, ny, {\d o}, \.{o}, sh, {\d u}& yes, 2 tones  & yes & SVO & agglutinative & little affixation & vestigial  \\
    Kinyarwanda (kin) & 30 & q, x & cy, jy, nk, nt, ny, sh & yes, 2 tones & no & SVO & agglutinative & strong prefixing & active, 16 \\
    Luganda (lug) & 25 & h, q, x & \textipa{\ng}, ny & yes, 3 tones & no & SVO & agglutinative & strong prefixing & active, 20 \\
    Luo (luo) & 31 & c, q, x, v, z &   ch, dh, mb, nd, ng’, ng, ny, nj, th, sh & yes, 4 tones & no & SVO  & agglutinative & equal prefixing and suffixing & absent \\
    Mossi (mos) & 26 & c, j, q, x  & ', \textepsilon, \textiota, \textscriptv & yes, 2 tones  & yes  & SVO  & isolating & strongly suffixing & active, 11 \\
    Chichewa (nya) & 31 & q, x, y & ch, kh, ng, \textipa{\ng}, ph, tch, th, \^{w} & yes, 2 tones & no  & SVO  & agglutinative & strong prefixing & active, 17\\
    Naija (pcm) & 26 & -- & -- & no  & no & SVO & mostly analytic & strongly suffixing & absent\\
    \shona (sna) & 29 & c, l, q, x & bh, ch, dh, nh, sh, vh, zh  &  yes, 2 tones & no & SVO & agglutinative & strong prefixing & active, 20 \\
    Swahili (swa) & 33 & x, q & ch, dh, gh, kh, ng', ny, sh, th, ts & no & yes   & SVO  & agglutinative & strong suffixing & active, 18 \\
    Setswana (tsn) & 36 & c, q, v, x, z & \^{e}, kg, kh, ng, ny, \^o, ph, \v{s}, th, tl, tlh, ts, tsh, t\v{s}, t\v{s}h & yes, 2 tones & no & SVO & agglutinative & strong prefixing & active, 18\\
    Akan/Twi (twi) & 22 & c,j,q,v,x,z & \textepsilon, \textopeno & yes, 5 tones  & no & SVO & isolating & strong prefixing &  active, 6\\
    Wolof (wol) &29& h,v,z & \textipa{\ng}, \`a, \'e, \"{e}, \'o, \~{n}  & no & yes  & SVO & agglutinative & strong suffixing & active, 10 \\
    \xhosa (xho) & 68 & -- & bh, ch, dl, dy, dz, gc, gq, gr, gx, hh, hl, kh, kr, lh, mh, ng, ngc, ngh, ngq, ngx, nkq, nkx, nh, nkc, nx, ny, nyh, ph, qh, rh, sh, th, ths, thsh, ts, tsh, ty, tyh, wh, xh, yh, zh &  yes, 2 tones & no & SVO & agglutinative & strong prefixing & active, 17\\
    \yoruba (yor) & 25 & c, q, v, x, z & {\d e}, gb, {\d s} , {\d o} & yes, 3 tones & yes  & SVO & isolating & little affixation & vestigial, 2 \\
    \zulu (zul) & 55 & -- & nx, ts, nq, ph, hh, ny, gq, hl, bh, nj, ch, ngc, ngq, th, ngx, kl, ntsh, sh, kh, tsh, ng, nk, gx, xh, gc, mb, dl, nc, qh &  yes, 3 tones & no & SVO  & agglutinative & strong prefixing & active, 17 \\
    \bottomrule
  \end{tabular}
  }
  \vspace{-3mm}
  \caption{Linguistic Characteristics of the Languages}
  \label{tab:lang_char}
  \end{center}
\end{table*}

\begin{table*}[tb]
\small
\centering
 \setlength\tabcolsep{4pt}
\begin{tabular}{lrr||lrr}
\toprule
& \textbf{No. agreed} & \textbf{agreed} & & \textbf{No. agreed} & \textbf{agreed} \\
\textbf{Lang.} & \textbf{annotation} & \textbf{annotation (\%)} & \textbf{Lang.} & \textbf{annotation} & \textbf{annotation (\%)}\\
\midrule
bam               & 1,091 &  77.9  & pcm & 1,073 & 76.6 \\
ewe               & 616 &  44.0  & tsn & 1,058 & 24.4 \\
hau               & 1,079 &  77.1  & twi & 1,306 & 93.2 \\
kin               & 1,127 &  80.5  & xho & 1,378 & 98.4 \\
lug               & 937 & 66.9   & yor & 1,059 & 75.6 \\
luo               & 564 & 40.3   & zul & 905 & 64.6 \\
mos               & 829 & 49.2   &  &  & \\
\bottomrule

\end{tabular}
\caption{\textbf{Number of sentences with agreed annotations and their percentages}}  
\label{tab:agreement}
\end{table*}

\begin{table}[th!]
 \footnotesize
 \begin{center}
\scalebox{0.82}{
  \begin{tabular}{llr}
    \toprule
    \textbf{Language} & \textbf{Data Source} & \textbf{\# Train}/\textbf{\# dev}/ \textbf{\# test} \\
    \midrule
    Afrikaans (\texttt{afr}) & UD\textunderscore{Afrikaans-AfriBooms} & 1,315/ 194/ 425 \\
    Arabic (\texttt{ara}) & UD\textunderscore{Arabic-PADT} & 6,075/ 909/ 680 \\
    English (\texttt{eng})  & UD\textunderscore{English-EWT} & 12,544/ 2001/ 2077 \\
    French (\texttt{fra}) & UD\textunderscore{French-GSD}  & 14,450/ 1,476/ 416  \\
    Naija (\texttt{pcm}) & UD\textunderscore{Naija-NSC} & 7,279/ 991/ 972  \\
    Romanian (\texttt{ron})  & UD\textunderscore{Romanian-RRT}  & 8,043/ 752/ 729  \\
    Wolof (\texttt{wol}) & UD\textunderscore{Wolof-WTB} & 1,188/ 449/ 470 \\
    \bottomrule
  \end{tabular}
 }
  \vspace{-3mm}
  \caption{\textbf{Data Splits for UD POS datasets} used as source languages for cross-lingual transfer. }
  \vspace{-4mm}
  \label{tab:ud_pos_data_split}
  \end{center}
\end{table}

\begin{table*}[ht]
\footnotesize
 \begin{center}
\begin{tabular}{llrr}

\toprule
\textbf{ Language} & \textbf{Source} & \textbf{Size (MB)}  \\
\midrule
Bambara (bam) & MAFAND-MT~\cite{adelani-etal-2022-thousand} & 0.8MB \\
\ghomala (bbj) & MAFAND-MT~\cite{adelani-etal-2022-thousand} & 0.4MB\\
\ewe (ewe) & MAFAND-MT~\cite{adelani-etal-2022-thousand} &  0.5MB \\
Fon (fon) & MAFAND-MT~\cite{adelani-etal-2022-thousand} &  1.0MB \\
Hausa (hau) & VOA~\cite{palen-michel-etal-2022-multilingual} & 46.1MB \\
Igbo (ibo) & BBC Igbo~\citep{ogueji-etal-2021-small} & 16.6MB \\
Kinyarwanda (kin) & KINNEWS~\cite{niyongabo-etal-2020-kinnews} & 35.8MB \\
Luganda (lug) & Bukedde~\cite{alabi-etal-2022-adapting} & 7.9MB \\
Luo (luo) & Ramogi FM news~\cite{adelani-etal-2021-masakhaner} and MAFAND-MT~\cite{adelani-etal-2022-thousand} & 1.4MB  \\
Mossi (mos) & MAFAND-MT~\cite{adelani-etal-2022-thousand} & 0.7MB \\
Naija (pcm) & BBC~\cite{alabi-etal-2022-adapting} & 50.2MB \\
Chichewa (nya) & Nation Online Malawi~\cite{Siminyu2021AI4DA} &  4.5MB \\
\shona (sna) & VOA~\cite{palen-michel-etal-2022-multilingual} &  28.5MB\\
Kiswahili (swa) & VOA~\cite{palen-michel-etal-2022-multilingual} & 17.1MB \\
Setswana (tsn) & Daily News~\cite{adelani-etal-2021-masakhaner}, MAFAND-MT~\cite{adelani-etal-2022-thousand} &  1.9MB \\
Twi (twi) & MAFAND-MT~\cite{adelani-etal-2022-thousand} &  0.8KB \\
Wolof (wol) & Lu Defu Waxu, Saabal, Wolof Online, and MAFAND-MT~\cite{adelani-etal-2022-thousand}  & 2.3MB  \\
\xhosa (xho) & Isolezwe Newspaper & 17.3MB \\
\yoruba (yor) & BBC \yoruba~\cite{alabi-etal-2022-adapting}  & 15.0MB \\
\zulu (zul) & Isolezwe Newspaper &  34.3MB \\
\midrule
Romanian (ron) & Wikipedia &  500MB \\
French (fra) & Wikipedia (a subset) &  500MB \\
\bottomrule 

\end{tabular}
\footnotesize
  \caption{Monolingual News Corpora used for language adapter and SFT training, and their sources and size (MB)}
  \label{tab:news_corpus}
\end{center}
\end{table*}

\begin{table*}[t]
\begin{center}
\footnotesize
\resizebox{\textwidth}{!}{%
\begin{tabular}{lraaarabbrabrbbbaabab|cc}
\toprule
\textbf{Method} & \textbf{bam} & \textbf{bbj} & \textbf{ewe} & \textbf{fon} & \textbf{hau} & \textbf{ibo} & \textbf{kin} & \textbf{lug} & \textbf{luo} & \textbf{mos} & \textbf{nya} & \textbf{pcm} & \textbf{sna} & \textbf{swa} & \textbf{tsn} & \textbf{twi}  & \textbf{wol} & \textbf{xho} & \textbf{yor} & \textbf{zul} & \textbf{AVG} & \textbf{AVG*}  \\
\midrule
\multicolumn{7}{l}{\texttt{\textbf{ara}} \textbf{as a source language}} \\
FT-Eval & 26.4 & 10.0 & 16.0 & 14.2 & 47.7 & 62.5 & 57.1 & 35.4 & 15.3 & 17.0 & 53.7 & 66.4 & 56.0 & 58.4 & 42.9 & 14.1 & 13.5 & 39.0 & 46.9 & 44.8 & 36.9 & 37.1 \\
LT-SFT & 41.0 & 30.7 & 41.2 & 45.0 & 47.3 & 62.9 & 54.0 & 48.7 & 56.2 & 43.2 & 54.4 & 63.3 & 53.6 & 59.4 & 44.8 & 39.9 & 51.0 & 36.8 & 50.6 & 44.8 & 48.4 & 48.0 \\
MAD-X & 44.5 & 36.5 & 50.9 & 45.9 & 48.5 & 59.5 & 55.5 & 51.1 & 60.5 & 46.7 & 53.4 & 66.8 & 53.8 & 59.1 & 40.4 & 37.9 & 52.3 & 40.3 & 52.3 & 44.6 & 50.0 & 49.7\\ 
\midrule
\multicolumn{7}{l}{\texttt{\textbf{pcm}} \textbf{as a source language}} \\
FT-Eval & 16.0 & 8.6 & 14.3 & 4.9 & 58.0 & 64.9 & 48.9 & 35.9 & 13.0 & 11.0 & 47.5 & 74.6 & 51.9 & 50.9 & 32.8 & 5.3 & 7.3 & 25.9 & 46.9 & 30.9 & 32.8 & 33.2 \\
LT-SFT & 44.4 & 39.4 & 51.1 & 38.1 & 59.2 & 66.6 & 47.9 & 53.5 & 61.3 & 52.3 & 49.3 & 75.3 & 48.9 & 50.6 & 40.8 & 35.3 & 63.9 & 25.1 & 58.3 & 30.6 & 49.6 & 48.8 \\
MAD-X & 42.1 & 43.6 & 53.5 & 39.4 & 57.3 & 68.2 & 55.7 & 58.1 & 60.1 & 51.9 & 59.6 & 75.8 & 57.5 & 55.7 & 44.8 & 36.9 & 58.9 & 32.9 & 57.1 & 40.6 & 52.5 & 51.8 \\ 
\midrule
\multicolumn{7}{l}{\texttt{\textbf{afr}} \textbf{as a source language}} \\
FT-Eval & 54.8 & 25.4 & 38.3 & 31.3 & 61.4 & 73.6 & 67.1 & 48.6 & 29.4 & 35.2 & 56.1 & 77.3 & 56.0 & 57.5 & 49.0 & 32.9 & 32.5 & 43.8 & 63.8 & 44.3 & 48.9 & 49.4 \\
LT-SFT & 69.2 & 55.6 & 64.0 & 52.5 & 62.8 & 74.7 & 66.1 & 59.0 & 69.4 & 63.4 & 54.4 & 79.7 & 58.4 & 57.1 & 48.5 & 49.0 & 79.3 & 41.0 & 64.3 & 41.5 & 60.5 & 59.6 \\
MAD-X & 61.9 & 56.1 & 63.9 & 53.0 & 63.0 & 75.2 & 68.2 & 60.2 & 68.1 & 63.4 & 62.0 & 80.8 & 61.1 & 60.6 & 50.4 & 48.6 & 75.7 & 43.8 & 65.2 & 46.0 & 61.4 & 60.6 \\ 
\midrule

\multicolumn{7}{l}{\texttt{\textbf{fra}} \textbf{as a source language}} \\
FT-Eval & 41.0 & 15.2 & 27.5 & 16.1 & 64.1 & 73.0 & 67.7 & 53.4 & 21.9 & 21.3 & 65.2 & 77.9 & 64.4 & 62.2 & 51.8 & 16.8 & 17.7 & 45.8 & 61.6 & 46.5 & 45.6 & 46.1 \\
LT-SFT & 60.6 & 52.2 & 63.3 & 60.2 & 63.9 & 75.6 & 63.4 & 57.6 & 69.0 & 65.2 & 66.4 & 79.7 & 63.0 & 61.2 & 52.4 & 48.6 & 78.3 & 43.9 & 64.7 & 44.3 & 61.7 & 60.7 \\
MAD-X & 62.0 & 57.9 & 64.2 & 59.4 & 66.9 & 78.7 & 71.3 & 64.1 & 74.0 & 67.7 & 70.2 & 83.4 & 68.6 & 65.4 & 53.0 & 48.1 & 78.3 & 46.0 & 67.8 & 50.2 & 64.9 & 63.9 \\ 
\midrule
\multicolumn{7}{l}{\texttt{\textbf{eng}} \textbf{as a source language}} \\
FT-Eval & 52.1 & 31.9 & 47.8 & 32.5 & 67.1 & 74.5 & 63.9 & 57.8 & 38.4 & 45.3 & 59.0 & 82.1 & 63.7 & 56.9 & 52.6 & 35.9 & 35.9 & 45.9 & 63.3 & 48.8 & 52.6 & 52.9 \\
LT-SFT & \textbf{67.9} & 57.6 & 67.9 & 55.5 & 69.0 & 76.3 & 64.2 & 61.0 & 74.5 & 70.3 & 59.4 & 82.4 & 64.6 & 56.9 & 49.5 & 52.1 & 78.2 & 45.9 & 65.3 & 49.8 & 63.4 & 62.5 \\
MAD-X & 62.9 & 58.5 & 68.7 & 55.8 & 67.0 & 77.8 & 70.9 & 65.7 & 73.0 & 71.8 & 70.1 & 83.2 & 69.8 & 61.2 & 49.8 & 53.0 & 75.2 & 57.1 & 66.9 & 60.9 & 66.0 & 65.2 \\ 
 \midrule
\multicolumn{7}{l}{\texttt{\textbf{ron}} \textbf{as a source language}} \\
FT-Eval & 46.5 & 30.5 & 37.6 & 30.9 & 67.3 & 77.7 & 73.3 & 56.9 & 36.7 & 40.6 & 62.2 & 78.9 & 66.3 & 61.0 & 55.8 & 35.7 & 33.8 & 49.6 & 63.5 & 56.3 & 53.1 & 53.4 \\
LT-SFT & 60.6 & 57.0 & 64.9 & 60.4 & 67.5 & 77.4 & 68.2 & 58.5 & 70.2 & 67.9 & 58.2 & 78.1 & 64.6 & 59.7 & 57.4 & 55.7 & 81.9 & 46.3 & 64.8 & 51.2 & 63.5 & 62.4\\ 
MAD-X & 63.5 & 62.2 & 66.6 & 61.8 & 66.5 & 80.0 & 73.5 & 62.7 & 76.5 & 71.8 & 66.0 & 83.7 & 71.1 & 64.5 & \textbf{61.2} & 53.5 & 79.5 & 48.6 & 69.5 & 57.8 & 67.0 & 66.1  \\
\midrule
\multicolumn{7}{l}{\texttt{\textbf{wol}} \textbf{as a source language}} \\
FT-Eval & 40.8 & 36.5 & 39.8 & 37.4 & 55.1 & 58.6 & 49.2 & 51.8 & 35.1 & 44.9 & 49.0 & 51.6 & 53.8 & 42.9 & 45.0 & 38.4 & 88.6 & 46.0 & 52.5 & 45.5 & 48.1 & 45.6 \\
LT-SFT (N) & 64.4 & 64.3 & 69.8 & 63.0 & 67.0 & 79.7 & 63.7 & 64.0 & 74.1 & 72.2 & 56.5 & 72.7 & 67.7 & 53.0 & 51.3 & 56.2 & 92.5 & 46.0 & 69.8 & 47.7 & 64.8 & 63.1 \\ 
MAD-X (N) & 46.6 & 41.8 & 47.2 & 37.8 & 53.9 & 51.8 & 41.0 & 39.0 & 46.5 & 44.0 & 38.3 & 40.2 & 44.3 & 38.8 & 44.6 & 40.1 & 85.6 & 39.2 & 46.4 & 45.2 & 43.0
 & 43.3 \\ 
MAD-X (N+W) & 61.7 & 63.6 & 68.9 & 63.1 & 66.8 & 77.0 & 67.8 & 69.1 & 73.7 & 71.3 & 63.2 & 75.1 & 68.9 & 55.8 & 50.7 & 54.9 & 90.4 & 49.6 & 70.0 & 51.7 & 65.7 & 64.1 \\ 
\midrule
\multicolumn{7}{l}{\texttt{\textbf{sna}} \textbf{as a source language}} \\
FT-Eval & 42.6 & 26.2 & 41.7 & 29.5 & 60.5 & 68.2 & 73.7 & 75.0 & 42.2 & 34.9 & 69.3 & 65.7 & 89.2 & 63.4 & 48.9 & 33.3 & 35.8 & 59.5 & 59.2 & 67.9 & 54.3 & 53.4 \\
LT-SFT & 52.2 & 57.5 & 66.0 & 55.4 & 60.5 & 71.9 & 69.0 & 80.1 & 75.7 & 58.1 & 70.4 & 60.2 & \textbf{89.9} & 63.5 & 50.6 & \textbf{65.8} & 71.6 & \textbf{62.7} & 62.2 & \textbf{72.9} & 65.8 & 64.2 \\
MAD-X & 50.3 & 57.0 & 65.3 & 56.3 & 64.1 & 71.9 & \textbf{75.0} & 79.2 & 75.9 & 59.8 & 70.6 & 68.6 & 89.7 & 63.2 & 52.7 & 61.0 & 75.3 & 61.8 & 57.8 & 69.8 & 66.3 & 64.5 \\ 
\midrule
\multicolumn{7}{l}{\texttt{\textbf{multi-source: eng-ron-wol}}} \\
FT-Eval & 44.2 & 36.3 & 39.3 & 39.3 & 69.4 & 78.5 & 70.6 & 59.2 & 35.5 & 46.8 & 60.9 & 81.4 & 65.8 & 58.5 & 53.8 & 38.8 & 89.1 & 48.8 & 65.2 & 53.5 & 56.7 & 54.4 \\
LT-SFT & 67.4 & 64.6 & 70.0 & 64.2 & 70.4 & 81.1 & 68.7 & 63.9 & 76.4 & \textbf{73.9} & 58.8 & 83.0 & 69.6 & 57.3 & 52.7 & 57.2 & 93.1 & 45.8 & 69.8 & 48.3 & 66.8 & 65.2\\ 
MAD-X & 66.2 & \textbf{65.5} & 70.3 & 64.9 & 69.1 & 82.3 & 73.1 & 68.0 & 75.1 & 74.2 & 69.2 & 83.9 & 69.4 & 62.6 & 53.6 & 55.2 & 90.1 & 52.3 & 70.8 & 59.4 & 68.8 & 67.5 \\ 
\midrule
\multicolumn{7}{l}{\texttt{\textbf{multi-source: eng-ron-wol-sna}}} \\
FT-Eval & 45.1 & 35.9 & 39.6 & 41.0 & 69.5 & 78.7 & 76.9 & 71.7 & 37.4 & 46.8 & 71.9 & 82.4 & 88.9 & 63.8 & 51.7 & 38.8 & 89.2 & 59.6 & 65.6 & 67.3 & 61.1 & 58.0 \\
LT-SFT & 66.7 & 64.7 & 68.5 & \textbf{65.1} & \textbf{71.0} & 81.2 & 75.3 & 80.2 & \textbf{79.3} & 73.5 & 73.6 & 83.6 & 89.1 & 64.3 & 51.1 & 60.9 & \textbf{93.2} & 61.8 & 69.1 & 70.2 &  \textbf{72.1} & \textbf{70.0}\\ 
MAD-X & 59.0 & 64.3 & \textbf{70.9} & 64.3 & 69.8 & \textbf{82.5} & 76.9 & \textbf{80.9} & 78.8 & 70.1 & \textbf{74.2} & \textbf{85.1} & 89.1 & \textbf{65.7} & 55.0 & 60.7 & 86.5 & 60.7 & \textbf{71.0} & 69.6 & 71.8 & \textbf{70.0} \\ 
\bottomrule
    \end{tabular}
}
\caption{\textbf{Cross-lingual transfer to MasakhaPOS }. Zero-shot Evaluation using FT-Eval, LT-SFT, and MAD-X, with \texttt{ron}, \texttt{eng}, \texttt{wol} and \texttt{sna} as source languages. Experiments are based on AfroXLMR-base. Non-Bantu Niger-Congo languages highlighted with \colorbox{Gray}{gray} (except for Bambara that is often disputed as a different language family --- Mande) while those of Bantu Niger-Congo languages are highlighted with \colorbox{LightCyan}{cyan} . AVG* excludes \texttt{sna} and \texttt{wol} from the average since they are source languages. }
\label{tab:cross-lingual_transfer_all}
  \end{center}
\end{table*}

\begin{figure*}[t]
    \centering
    \includegraphics[width=\linewidth]{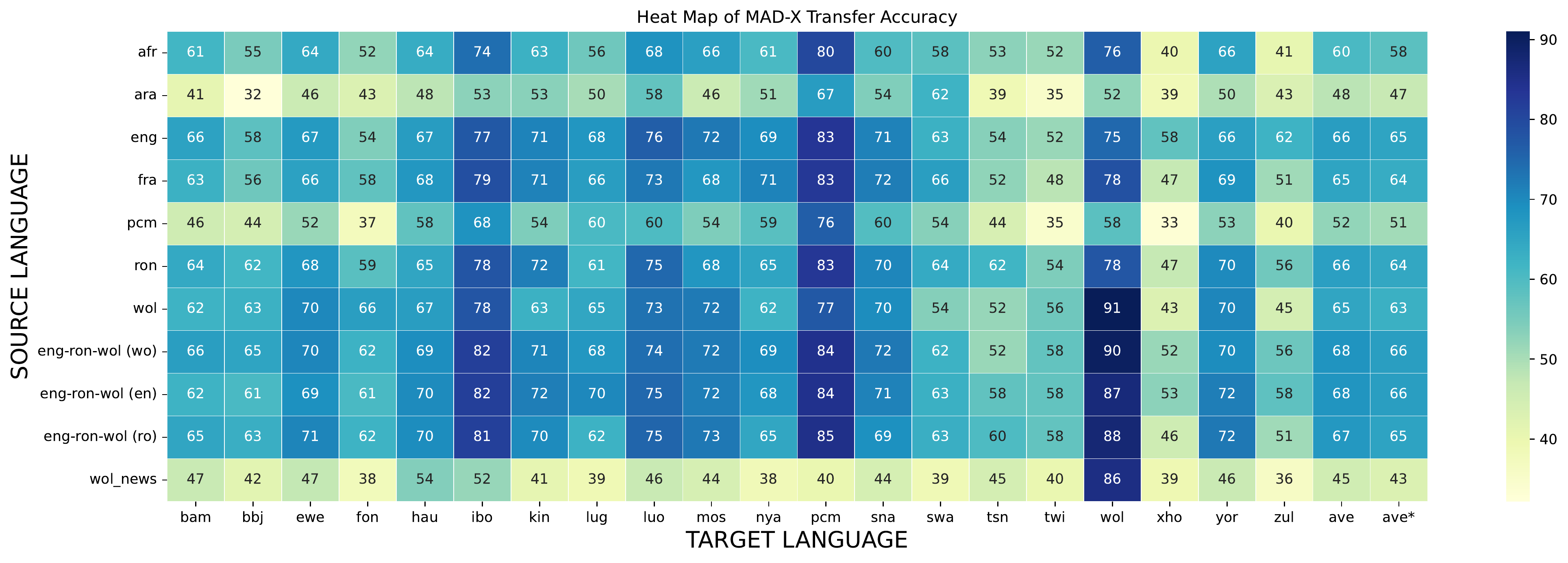}
    \vspace{-7mm}
    \caption{\textbf{MAD-X: Cross-lingual Experiments on MasakhaPOS }. Zero-shot Evaluation using \texttt{afr}, \texttt{ara}, \texttt{eng}, \texttt{fra}, \texttt{ron}, \texttt{pcm} and \texttt{wol} as source languages. Experiments based on AfroXLMR-base. \textbf{ave*} excludes pcm and wol from the average since they are also source languages. }
    \label{fig:transfer_madx}
    \vspace{-4mm}
\end{figure*}
\section{Language Characteristics}
\label{sec:appendix_lang_char}
\autoref{tab:lang_char} provides the details about the language characteristics. 

\section{Annotation Agreement}
\label{sec:annotation_agreement}
\autoref{tab:agreement} provides POS annotation agreements at the sentence level for 13 out of the 20 focus languages.

\section{UD POS data split}
\label{sec:ud_pos_data}
\autoref{tab:ud_pos_data_split} provides the UD POS corpus found online that we make use for determining the best transfer languages

\section{Hyper-parameters for Experiments}
\label{sec:parameters}
\paragraph{Hyper-parameters for Baseline Models}
The PLMs were trained for 20 epochs with a learning rate of 5e-5 using huggingface transformers \citep{wolf-etal-2020-transformers}. We make use of a batch size of 16

\paragraph{Hyper-parameters for adapters} We train the task adapter using the following hyper-parameters: batch size of 8, 20 epochs, ``pfeiffer'' adapter config, adapter reduction factor of 4 (except for Wolof, where we make use of adapter reduction factor of 1), and learning rate of 5e-5. For the language adapters, we make use of 100 epochs or maximum steps of 100K, minimum number of steps is 30K, batch size of 8, ``pfeiffer+inv'' adapter config, adapter reduction factor of 2, learning rate of 5e-5, and maximum sequence length of 256. 

\paragraph{Hyper-parameters for LT-SFT} We make use of the default setting used by the \citet{ansell-etal-2022-composable} paper. 

\section{Monolingual data for Adapter/SFTs language adaptation}
\label{sec:news_corpora}
\autoref{tab:news_corpus} provides the UD POS corpus found online that we make use for determining the best transfer languages

\section{MAD-X multi-source fine-tuning}
\label{sec:madx_multisource}
\autoref{fig:transfer_madx} provides the result of MAD-X with different source languages, and multi-source fine-tuning using either \texttt{eng}, \texttt{ron} or \texttt{wol} as language adapter for task adaptation prior to zero-shot transfer. Our result shows that making of \texttt{wol} as language adapters leads to slightly better accuracy (69.1\%) over \texttt{eng} (68.7\%) and \texttt{ron} (67.8\%). But in general, either one can be used, and they all give an impressive performance over LT-SFT, as shown in \autoref{tab:cross-lingual_transfer_all}.

\section{Cross-lingual transfer from all source languages}
\label{sec:cross_lingual_all}
\autoref{tab:cross-lingual_transfer_all} shows the result of cross-lingual transfer from each source language (\texttt{afr}, \texttt{ara}, \texttt{eng}, \texttt{fra}, \texttt{pcm}, \texttt{ron}, and \texttt{wol}) to each of the African languages.  We extended the evaluation to include \texttt{sna} (since it was recommended as the best transfer language for a related task -- named entity recognition by \cite{adelani-etal-2022-masakhaner}) by using the newly created POS corpus. We also tried other Bantu languages like \texttt{kin} and \texttt{swa}, but their performance was worse than \texttt{sna}.  Our evaluation shows that \texttt{sna} results in better transfer to Bantu languages because of it's rich morphology. We achieved the best result for all languages using multi-source transfer from (\texttt{eng}, \texttt{ron}, \texttt{wol}, \texttt{sna}) languages. 

\end{document}